%% file: main.tex
\documentclass{article} 
\usepackage{iclr2022_conference,times}
\usepackage{graphicx}
\usepackage{times}
\usepackage{epsfig}
\usepackage{amsmath}
\usepackage{tabularx}
\usepackage{amssymb}
\usepackage{caption}
\usepackage{subcaption}
\usepackage{rotating}
\usepackage{ntheorem}
\usepackage{multirow}
\usepackage{hyperref}
\usepackage{hhline}
\usepackage{array}

\input{math_commands.tex}

\makeatletter
\renewtheoremstyle{plain}
    {\item{\theorem@headerfont ##1\ ##2\theorem@separator}~}
    {\item{\theorem@headerfont ##1\ ##2\ (##3)\theorem@separator}~}

\usepackage{hyperref}
\usepackage{url}

\newcommand\Tstrut{\rule{0pt}{2.6ex}}        
\newcommand\TstrutS{\rule{0pt}{2ex}}         
\newcommand\Bstrut{\rule[-0.9ex]{0pt}{0pt}}

\title{Physics Informed Convex Artificial Neural Networks (PICANNs) for Optimal Transport based Density Estimation}

\author{Amanpreet Singh,
Sarang Joshi

\thanks{A.~Singh and S.Joshi were supported by NSF grant DMS-1912030} \\

University of Utah\\ 
Salt Lake City, UT 84112, USA \\

\And

Martin Bauer
\thanks{M.~Bauer was supported by NSF grants DMS-1912037 and DMS-1953244.}\\
Florida State University\\
Tallahassee, FL 32306, USA \\
}

\begin{document}

\newtheorem{thm}{Theorem}[section]
\newtheorem{dfn}[thm]{Definition}
\numberwithin{thm}{section} 

\maketitle
\begin{abstract}
  Optimal Mass Transport (OMT) is a well-studied problem with a variety of applications in a  diverse set of fields, ranging from Physics to Computer Vision and in particular Statistics and Data Science. Since the original formulation of Monge in 1781 significant theoretical progress been made on the existence, uniqueness and properties of the optimal transport maps. The actual numerical computation of the transport maps, particularly in high dimensions, remains a challenging problem. In the past decade several neural network based algorithms have been proposed to tackle this task. In this paper, building on recent developments of input convex neural networks and physics informed neural networks for solving PDE's, we propose a new Deep Learning approach to solve the continuous OMT problem. Our framework is based on Brenier's theorem, which reduces the continuous OMT problem to that of solving a nonlinear PDE of Monge-Ampere type whose solution is a convex function. 
  To demonstrate the accuracy of our framework we compare our method to several other deep learning based algorithms.
  We then
  focus on applications to the  ubiquitous density estimation and generative modeling tasks in statistics and machine learning. Finally as an example we present how our framework can be incorporated with an autoencoder to estimate an effective probabilistic generative model.
\end{abstract}

\section{Introduction}
Optimal Mass Transport (OMT) is a well-studied problem with a variety of applications in a diverse set of fields, ranging from physics to computer vision and in particular statistics and data science.
Fueled by the appearance of OMT in transformation-based density estimation and random sampling algorithms in machine learning applications, several new numerical frameworks for solving this optimization problem (in high dimensions) have been recently proposed, see, e.g., \cite{korotin2021neural} and the references therein.
Our article adds to this expanding list by developing a new framework for the estimation of the $L^2$-optimal transport problem.
Our algorithm, which is based on Brenier's theorem, builds on recent developments of input convex neural networks and physics-informed neural networks for solving PDE's. Before we describe the contributions of the article in more detail, we will briefly summarize the motivation of our investigations and recent developments in the field. 

{\bf Density estimation and random sampling:}
Density estimation and random sampling are fundamental problems in machine learning and statistical inference. The density estimation problem is to estimate a smooth probability density based on a discrete finite set of observations. In traditional parametric density estimation techniques, we assume that the data is drawn from a known parametric family of distributions, and it only remains to best estimate these parameters. These methods require that we have a basis to believe that the data is indeed derived from a specific family of distributions and are consequently limited in their applicability to many modern tasks. One of the most ubiquitous parametric techniques is Gaussian Mixture Modeling~\citep{mclachlan1988mixture}.

Nonparametric techniques were first proposed by \cite{fix1951nonparametric} (\cite{silverman}) to move away from such rigid distributional assumptions. The most used approach is the kernel density estimation, which dates back to~\citet{rosenblatt} and~\citet{parzen}. 
Despite decades of work in this field, many challenges remain regarding the implementation and practical performance of kernel density estimators, including in particular, the bandwidth selection and the lack of local adaptivity resulting in a large sensitivity to outliers~\citep{loader}. These problems are particularly exacerbated in high dimensions with the curse of dimensionality.

Recently, diffeomorphic transformation-based algorithms have been proposed to tackle this problem~\citep{dinh2017,marzouk2016sampling,younes2020,bauer2017diffeomorphic}.  The basic concept of transformation-based algorithms is to find a diffeomorphic mapping between a reference probability distribution and the unknown target distribution, from which the data is drawn. Consequently, transformation-based density estimation leads at the same time to an efficient generative model, as new samples from the estimated density can be generated at a low cost by sampling from the reference density and transforming the samples by the estimated transformation. The fundamental problem in diffeomorphic transformation-based approaches is how to estimate and select the transformation: from a theoretical point of view there exists an infinite set of transformations that map two given probability densities onto each other. Recently, several deep learning methods have been devised for this task, where Normalizing Flows (NF) stand out among these methods. Examples of such models include Real NVP \citep{dinh2017}, Masked Autoregressive Flows ~\citep{papamakarios2017masked}, iResNets ~\citep{behrmann2019invertible}, Flow++ \citep{ho2019flow++} and Glow ~\citep{kingma2018glow}. For a review
  of the vast NF literature, we refer to the the overview article~\citep{kobyzev2020normalizing}. Although these methods have shown to perform well in density estimation applications, the interpretability of the obtained transformation is less clear, e.g. in Real NVP~\citep{dinh2017}, the solution selection is obtained by restricitng the transformations to the class of diffeomorphisms with triangular Jacobians that are easy to invert, which is closely related to the Knothe-Rosenblatt rearrangement~\citep{knothe1957contributions,rosenblatt1952remarks}. 

{\bf Optimal mass transport:}  Optimal mass transport, on the other hand, formulates the transport map selection as the minimizer of a cost function~\citep{villani2008optimal,villani2003topics}. The optimal transportation cost induces a metric structure, the Wasserstein metric, on the space of probability densities and is sometimes referred to as the Earth Mover’s Distance. This theory, which dates back to 1781, was originally formulated by the French mathematician Gaspard~\citet{monge1781memoire}.  The difficulty in applying this framework to the proposed density estimation problem lies in solving the corresponding optimization problem, which in a dimension greater than one is highly non trivial. The fully discrete OMT problem (optimal assignment problem) can be solved using linear programming and can be approximated by the Sinkhorn algorithm~\citep{cuturi,papadakis}. However, these algorithms do not lead to a continuous transformation map and thus can't be used for the proposed diffeomorphic density estimation and generative modelling. Previous algorithmic solutions for the continuous OMT problem include fluid mechanics-based approaches~\citep{benamou2000computational}, finite element or finite difference-based methods~\citep{benamou2010two,benamou2019minimal} and steepest descent-based energy minimization approaches~\citep{angenent2003minimizing,carlier2010knothe,loeper2005numerical}.

In recent years, several deep learning methods have been deployed for solving the OMT problem. In these methods, the OMT problem is typically embedded in the loss function for the neural network model. Recent work by~\cite{OTICNN} proposed to approximate the OMT map as the solution of min-max optimization using input convex neural networks (ICNN), see~\cite{amos}. The min-max nature of this algorithm arises from the need to train an ICNN to represent a convex function and the conjugate of the convex function. Building upon this approach, \citet{korotin2019wasserstein} imposed a cyclic regularisation that converts the min-max optimization problem to a standard minimization problem. The conjugate convex function composed of the convex function itself should be identity helps in avoiding the min-max optimization. 
This change results in a faster converging algorithm that scales well to higher dimensions and also prevents convergence to local saddle points and instabilities during training, as is the case in the min-max algorithm.

Another class of neural networks which have been proposed to solve OMT problems are Generative Adversarial Networks(GANs) \citep{goodfellow2014generative}. GANs are defined through a min-max game of two neural networks where one of the networks tries to generate new samples from a data distribution, while the other network judges whether these generated samples originate from the data population or not. Later, \citet{gulrajani2017improved} proposed using the Wasserstein-1 distance in GANs instead of the Jensen-Shannon divergence between the generated distribution and the data distribution as in the original formulation. They demonstrated that this new loss functions leads to better stability of the training of networks attributed to the Wasserstein metric being well defined even when the two distributions do not share the same support.


{\bf Contributions:}
In this paper, we propose a different deep learning-based framework to approximate the optimal transport maps. 
The approach we present relies on Brenier's celebrated theorem~\citep{brenier}, thereby reducing the optimal transport problem to that of solving a partial differential equation: a Monge-Ampere type equation. We frame this PDE in the recently 
developed paradigm of Physics Informed Neural Networks (PINNs)~\citep{raissi}. Similar to other deep learning-based algorithms, our framework directly inherits the dimensional scalability of neural networks~\citep{shin2020convergence}, which traditional finite element or finite difference methods for solving PDEs do not possess. Brenier's theorem further states that the optimal transport map is given by the gradient of a convex function- the Brenier potential. To incorporate this information in our PINN approach, we parameterize the Brenier potential using an ICNN, thereby guaranteeing its convexity. 

We test the accuracy of our OMT solver on numerous synthetic examples for which analytical solutions are known. Our experiments show that our algorithm indeed approximates the true solution well, even in high dimensions. To further quantify the performance of the new framework, we compare it to 
two other deep learning-based algorithms, for which we guided the selection by the results of the  recent benchmarking paper by \citet{korotin2021neural}, in which they evaluate the methods presented in ~\cite{seguy2017large,nhan2019threeplayer,taghvaei20192,OTICNN,liu2019wasserstein,mallasto2019q,korotin2019wasserstein}. We restricted our comparision  
to the algorithms of \citet{OTICNN} and \citet{korotin2019wasserstein}, as these two showed the best performance in this benchmark. Our results showed that the newly proposed method significantly outperforms these methods in terms of accuracy. 

As an explicit application of our solution of OMT, we focus on the  density estimation problem.
In synthetic examples, we show that we can estimate the true density based on a limited amount of samples. In the appendix we also demonstrate the generative power of our framework by combining it with a traditional autoencoder and applying it to the MNIST data set. 


In accordance with the best practices for reproducible research, we are providing an open-source version of the code, which is publicly available on \href{https://github.com/4m4npr33t/PICANNs}{github}.


\section{OMT using Deep Learning}
In this section, we will present our framework for solving the Optimal Mass Transport (OMT) problem. Our approach will combine methods of deep learning with 
the celebrated theorem of Brenier, which reduces the solution of the OMT problem to solving a Monge-Ampere type equation. To be more precise, we will 
tackle this problem by embedding the Monge-Ampere equation into the broadly applicable concept of Physics Informed Neural Networks. 
\subsection{Mathematical Background of OMT}
We start by  summarizing the mathematical background of OMT, including a description of Brenier's theorem. For more information we refer to the vast literature on OMT, see e.g., \cite{villani2003topics,villani2008optimal}.

Let $\Omega$ be a convex and bounded domain of $\mathbb{R}^n$ and let $dx$ denote the standard measure on $\mathbb{R}^n$.
For simplicity, we restrict our presentation to the set $\mathcal{P}(\Omega)$ of all absolutely continuous measures on $\Omega$, 
i.e., $\mathcal{P}(\Omega) \ni\mu=fdx$ with $f\in L^1(\Omega)$, such that $\int_{\Omega} f dx=1$. 
From here, on we will identify the measure $\mu$ with its density function $f$.

We aim to minimize the cost 
of transporting a density $\mu$ to a density $\nu$ using a (transport) map $T$, which leads to the so-called Monge Optimal Transport Problem. To keep the presentation as simple as possible, we will consider only the special case of a quadratic cost function.
\begin{dfn}[$L^2$-Monge Optimal Transport Problem] Given $\mu, \nu \in \mathcal{P}(\Omega)$, 
 minimize
 $$\mathbb{M}(T) = \int_{\Omega} \|x- T(x)\|^2 d\mu(x)$$
over all $\mu$-measureable maps $T: \Omega \to \Omega$ subject to $\nu = T_*\mu$.
We will call an optimal $T$ an optimal transport map. 
\end{dfn}
Here, the constraint is formulated in terms of the push forward action of a measurable map $T: \Omega\to \Omega$, which is defined via
\begin{equation}
T_*\mu(B)=\mu(T^{-1}(B)),  
\end{equation}
for every measurable set $A\subset \Omega$.
By a change of coordinates, the constraint $T_*\mu=T_*(fdx)=\nu=g dx$  can be thus reduced to the equation
\begin{equation}\label{eq:pushforward}
    f(x) = g(T(x))|\operatorname{det}(D T(x))|.
\end{equation}
The above equation can be also expressed via the pullback action as $\mu=T^*\nu$.

The existence of an optimal transport map is not always guaranteed. We will, however, see, that in our situation, i.e., for absolutely continuous measures, the existence and uniqueness is indeed guaranteed. First, we will introduce a more general formulation of the Monge problem, the  Kantorovich formulation of OMT.

Therefore, we define the space of all transport plans $\Pi(\mu, \nu)$, i.e., of all measures on the product space 
$\Omega\times \Omega$, such that the first marginal is $\mu$ and the second marginal is $\nu$. The OMT problem in the Kantorovich formulation then reads as: 
\begin{dfn}[$L^2$-Kantorovich's Optimal Transport Problem]
Given $\mu, \nu \in \mathcal{P}(\Omega)$, minimize
$$ 
 \mathbb{K}(\pi) = \int_{\Omega \times \Omega} \|x- y\|^2d\pi(x, y)
$$
over all $\pi \in \Pi(\mu,\nu)$.
\end{dfn}
Note that the $L^2$-Wasserstein metric $W_{2}(\mu,\nu)$ between $\mu$ and $\nu$ is defined as the infimum of $\mathbb{K}$.

We will now formulate Brenier's theorem, which guarantees the existence of an optimal transport map and will be the central building block of our algorithm:
\begin{thm}[\cite{brenier}] \label{thm:Brenier} Let $\mu, \nu \in \mathcal{P}(\Omega)$. Then there exists a unique optimal transport plan $\pi^* \in \Pi(f, g)$, which is given by $\pi^*(x, y) =(\operatorname{id}\times T)$
where $T=\nabla u$ is the gradient of a convex function $u$ that pushes $\mu$ forward to $\nu$, i.e., $(\nabla u)_*\mu= \nu$.  The inverse $T^{-1}$ is also given by the gradient of a convex function that is the Legendre transform of the convex function $u$.
\end{thm} 
Thus, Brenier's Theorem  guarantees the existence and the uniqueness of the optimal transport map of the OMT problem.
Consequently, we can determine this optimal transport map by solving for the function $u$ in the form of a Monge-Ampère equation:
\begin{equation}\label{eq:Monge-Ampere}
    \operatorname{det}(D^2(u)(x))\cdot g(\nabla u(x)) = f(x)
\end{equation}
where $D^2$ is the Hessian, $\mu=fdx$ and $\nu=gdx$. 
We obtain~\eqref{eq:Monge-Ampere} directly from~\eqref{eq:pushforward} using the constraint that $T=\nabla u$ as required by Brenier's theorem. We will also refer to this map as the Brenier map. This map is a diffeomorphism as it is a gradient of a strictly convex function.
Using methods of classical numerical analysis, Brenier's theorem has been used e.g. in~\cite{peyre2019computational} to obtain a numerical framework for the continous OMT problem. 
In the following section we will propose a new discretization to this problem, which will make use of recent advances in deep learning. 

\begin{figure*}
    \centering
    \includegraphics[width=\textwidth]{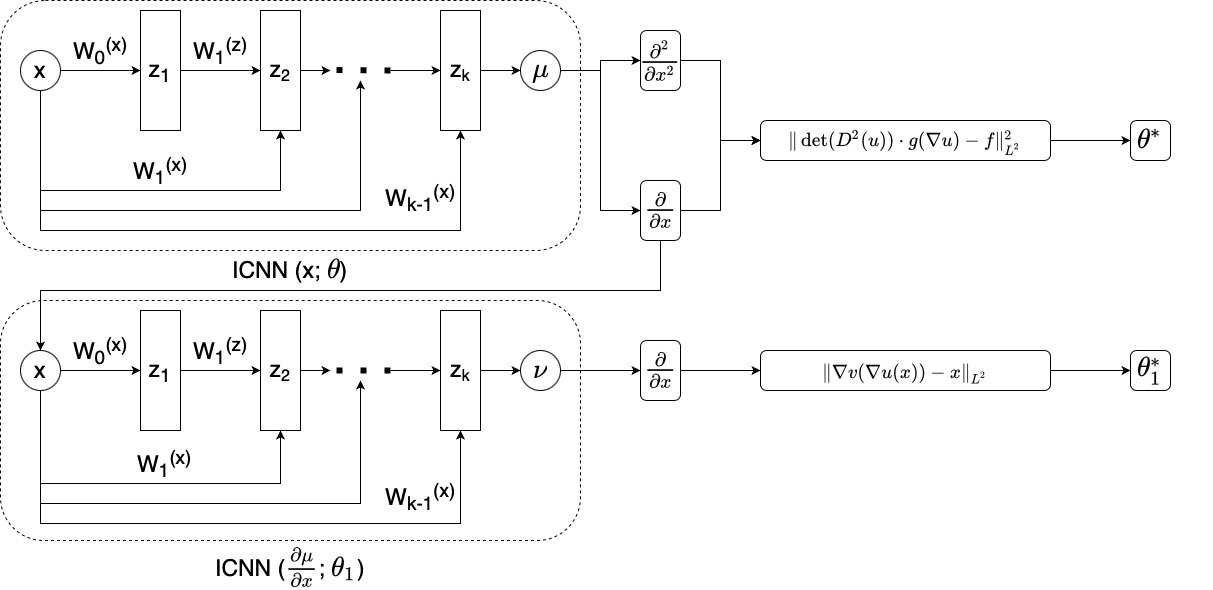}
    \caption{PICANN architecture. We present how a combination of two ICNN networks can be used to learn the forward and the inverse map between two distributions. Both these networks are trained independently with their respective loss functions. The inverse network uses the gradient of the output of the first network as its input.}
    \label{fig:PICANN_Architecture}
\end{figure*}

\subsection{Solving OMT using PINNs} \label{sec:OMTviaPINN}
Physics Informed Neural Networks (PINNs) were proposed by~\citet{raissi} to solve general nonlinear partial differential equations (PDEs). The basic concept is to use the universal approximation property of deep neural networks to represent the solution of a PDE via a network. Using
the automatic differentiation capability of modern machine learning frameworks, a loss function is formulated, such that its minimizer solves the PDE in a weak sense. Such a loss function encodes the structured information, which results in the amplification of the information content of the data the network sees~\citep{raissi}. This formulation of the PDE results in good generalization even when only few training examples are available. 

PINNs have found widespread applications in a short period of time since their introduction. These applications include a wide variety of PDEs, including the Navier-Stokes equation~\citep{jin}, nonlinear stochastic PDEs~\citep{zhang} or Allen Cahn PDEs \citep{mcclenny}. 

In this work, we propose to use the PINN approach to solve the Monge-Ampere equation, as presented in~\eqref{eq:Monge-Ampere}, and hence implicitly the OMT problem. This equation has been extensively studied and the properties of its solutions are well established. By Theorem~\ref{thm:Brenier}, we know that the solution is given by a convex function $u$. Recently, \cite{amos} proposed a new architecture of neural networks, Input Convex Neural Networks (ICNNs), that explicitly constrains the function approximated by the network to be convex.  Consequently, this architecture naturally lends itself to our proposed application, as it directly encodes Brenier's theorem.

In the ICNN architecture, the activation function is a nondecreasing convex function and the internal weights ($W_n^{(x)}$) are constrained to be non-negative; see Figure~\ref{fig:PICANN_Architecture} for a schematic description of this class of networks. This architecture is derived from  two simple facts:  non-negative sums of convex functions are also convex, and the composition of a convex and convex nondecreasing function is again convex.


In the following equation, we assume that we are given $\mu=fdx$ and $\nu =gdx$.
The loss function corresponding to~\eqref{eq:Monge-Ampere} is then given by
\begin{equation}
    \| \operatorname{det}(D^2(u))\cdot g(\nabla u) - f \|_{L^2}^2
\end{equation}
where  $u$ is expressed as the output of a ICNN of sufficient depth and width.
Once we have estimated the optimal transport map, the $L^2$-Wasserstein metric between $\mu$ and $\nu$ is given by
\begin{equation}\label{L2-loss}
    \int \|x-\nabla u(x)\|^2\; g(x) dx.
\end{equation}
We call this combination of the PINN approach with the ICNN structure, Physics  Informed  Convex  Artificial Neural Networks (PICANNs). 

In several applications, we are interested in computing the inverse transformation 
at the same time. By a duality argument, we know that this map is also given by the gradient of a convex function. Thus, we use  a second ICNN to compute the inverse optimal transport map ($\nabla v$) by solving the minimization problem: 
 \begin{equation}\label{InverseLoss}
     \| \nabla v (\nabla u (x))- x\|_{L^2},
 \end{equation}
where $\nabla u$ is the optimal transport map solving $(\nabla u)_*\mu=\nu$.

\subsection{Diffeomorphic Random Sampling and Density Estimation}
In many applications, such as the Bayesian estimation, we can evaluate the density rather easily but generating samples from a given density is not trivial. Traditional methods include Markov Chain Monte Carlo methods, e.g., the Metropolis Hastings algorithm~\citep{hastings1970monte}. An alternative idea is to use diffeomorphic density matching between the given density $\nu$  and a standard density $\mu$ from which samples can be drawn easily. Once we have calculated the transport map, standard samples are transformed by the push-forward diffeomorphism to generate samples from the target density $\nu$. This approach has been followed in several articles, where the optimal transport map selection was based on both, the Fisher-Rao metric~\citep{bauer2017diffeomorphic} and the Knothe–Rosenblatt rearrangement~\citep{marzouk2016sampling}. The efficient implementation of the present paper directly leads to an efficient random sampling algorithm in high dimensions.

We now recall the density estimation problem using the OMT framework. We are given samples $x_i$ drawn from an unknown density $\mu \in \mathcal{P}(\Omega)$ that we aim to estimate. The main idea of our algorithm is to represent 
the unknown density as the pullback via a (diffeomorphic) Brenier map $\nabla u$ of a given background density $\nu=gdx$, i.e., $(\nabla u)^*\nu=\mu$ or equivalently $(\nabla u)_*\mu=\nu$.

As we do not have an explicit target density, but only a finite number of samples, we need to find a replacement for the $L^2$-norm used in \eqref{L2-loss} to estimate the transport map $\nabla u$. We 
do this by 
maximizing the log-likelihood of the data with respect to the density $(\nabla u)^*\nu$:
\begin{equation}\label{eq:16}
     \frac{1}{N} \sum_{i} \log\left(\operatorname{det}(D^2(u(x_i)))\cdot g(\nabla u(x_i)) \right).
\end{equation}
Using our PINNs framework, we represent the convex function $u$ again via an ICNN, which serves as an implicit regularizer. 
This equation can be alternatively interpreted as minimizing
the empirical Kullback-Leibler divergence
between $\mu$ and the pullback of the background density $\nu$. 


To generate new samples from the estimated density, we use the inverse map to transform the samples from the background density $\nu$. We calculate the inverse map using a second neural network and explicit loss function given by~\eqref{InverseLoss}.

\begin{figure*}[!htb]
    \centering
    \begin{subfigure}[t]{0.22\textwidth}
        \centering
        \includegraphics[width=\textwidth]{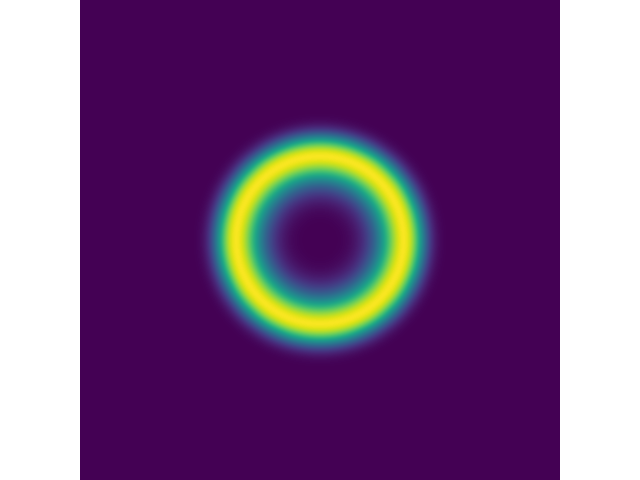}
        \caption{Ground truth}
        \label{subfig:True_Annulus}
    \end{subfigure}
    \begin{subfigure}[t]{0.22\textwidth}
        \centering
        \includegraphics[width=\textwidth]{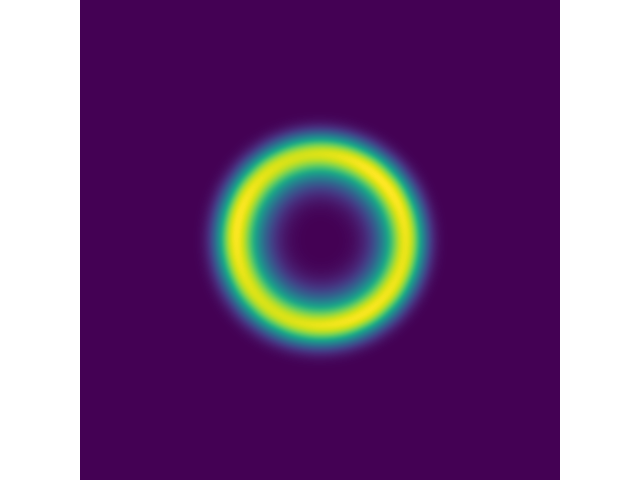}
        \caption{Est. pdf (PICANN)}
        \label{subfig:Estimated_Annulus}
    \end{subfigure}
    \vspace{.1cm}
    \begin{subfigure}[t]{0.22\textwidth}
        \centering
        \includegraphics[width=\textwidth]{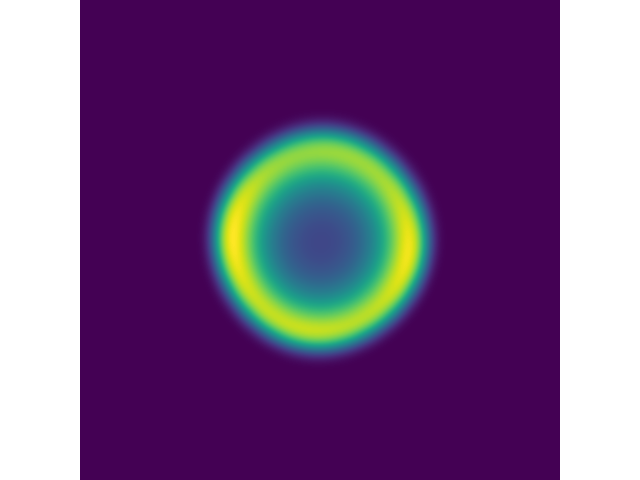}
        \caption{Est. pdf (W2 Gen)}
        \label{subfig:Diff}
    \end{subfigure}
    \begin{subfigure}[t]{0.22\textwidth}
        \centering
        \includegraphics[width=\textwidth]{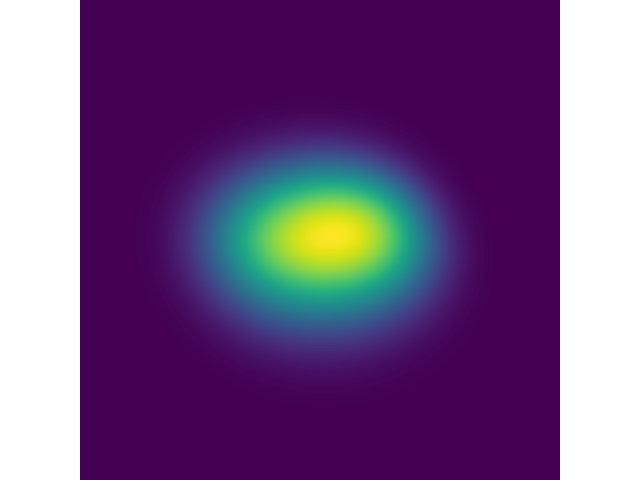}
        \caption{Est. pdf (OT ICNN)}
        \label{subfig:Diff}
    \end{subfigure}
    \begin{subfigure}[t]{0.22\textwidth}
        \centering
        \includegraphics[width=\textwidth]{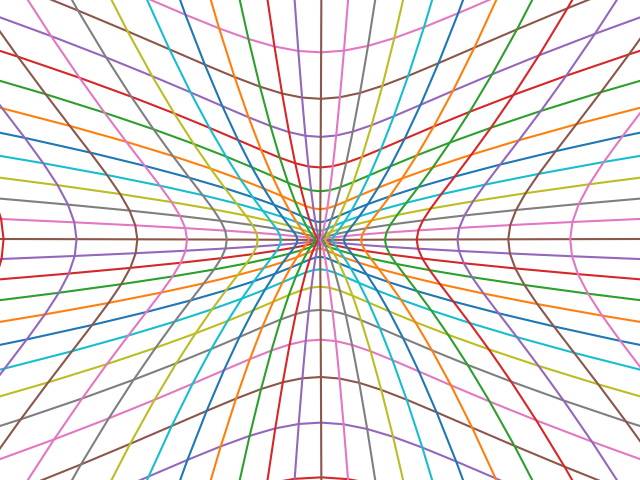}
        \caption{Ground truth}
        \label{subfig:True_Map}
    \end{subfigure}
    \begin{subfigure}[t]{0.22\textwidth}
        \centering
        \includegraphics[width=\textwidth]{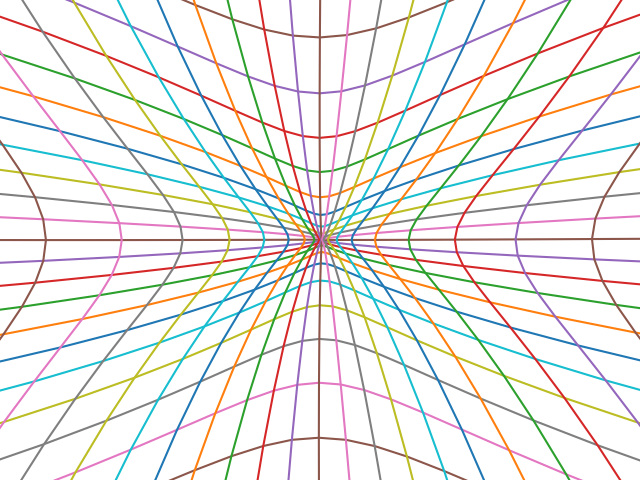}
        \caption{Est. map (PICANN)}
        \label{subfig:Estimated_Map}
    \end{subfigure}
    \begin{subfigure}[t]{0.22\textwidth}
        \centering
        \includegraphics[width=\textwidth]{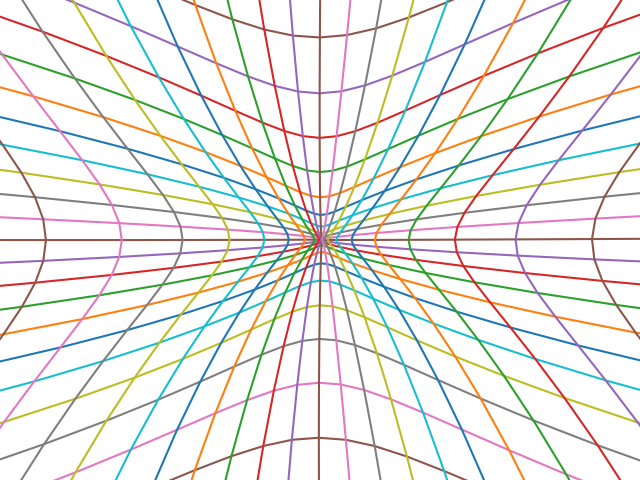}
        \caption{Est. map (W2 Gen)}
        \label{subfig:OT_ICNN_Estimated_Map}
    \end{subfigure}
    \begin{subfigure}[t]{0.22\textwidth}
        \centering
        \includegraphics[width=\textwidth]{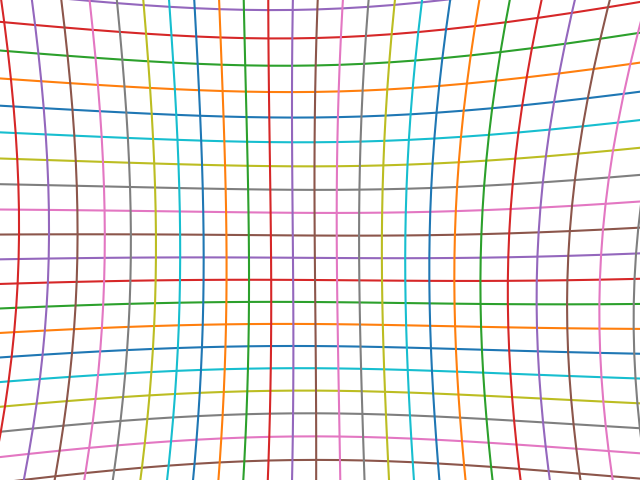}
        \caption{Est. map (OT ICNN)}
        \label{subfig:OT_ICNN_Estimated_Map}
    \end{subfigure}
    \caption{Validation: Panel~(a) shows the true annulus distribution, the estimated annulus distribution using the PICANN approach. The W2-GEN and OT-ICNN approaches are shown in Panels~(b), ~(c) and ~(d), respectively. Panel~(e) shows the analytical optimal transport map between the unit Gaussian and the annulus distribution. The estimated optimal transport map using the PICANN approach is presented in Panel~(f) and the maps estimated using the  W2-GEN and OT-ICNN methods are shown in Panels~(g) and ~(h), respectively.}
    \label{fig:Annulus_Results}
\end{figure*}

\section{Experimental Results}
In this section, we will detail our implementation and present several experiments demonstrating both the applicability and accuracy of our framework. In particular, we will compare our results in several experiments to state-of-the-art deep learning-based OMT solvers, and we will show that we outperform these methods in terms of accuracy.


\subsection{Network details} \label{sec:NetworkDetails}
As explained in Section~\ref{sec:OMTviaPINN}, we use an ICNN architecture for both the forward and the backward map in all of our experiments, c.f. Figure~\ref{fig:PICANN_Architecture}. As with every deep learning approach, we need to tune the hyperparameters, including width/depth of the network, activation functions and batch size. The width of the network needs to increase with the dimension of the ambient space of the data to ensure sufficient flexibility. For our experiments in lower dimensions, we used a network with three hidden layers with 128 neurons in each layer, whereas for experiments in 30d, we used a network with four hidden layers with 128 neurons in each layer. To initialize the network,  we first train the networks to learn the identity transformation, i.e., $\nabla u = I$, which we use as the initial starting point for all our experiments. In all our experiments, 10,000 target samples were used. 

To guarantee the convexity of the output function, the activation functions need to be convex and non-decreasing.  Since  simple ReLUs are not strictly convex and have a vanishing second derivative almost everywhere, we experimented with the family of Rectified Power Units (RePUs), the log exponential family and the 'Softplus' function. The Softplus function to the power of $\alpha$, which is defined via 
$
    \operatorname{Softplus}^{\alpha}(x) = \left(\log{\left(1 + \exp{x}\right)}\right)^{\alpha}
$, turned out to be best suited for our applications, where we chose $\alpha=1.1$. In particular our experiments suggested that networks with this activation function were able to generalize well to regions where no or only limited training data were available. 
\subsection{Validation and comparison to other methods}\label{sec:Valdidation}
To demonstrate the accuracy of our implementation, we present several experiments, in which analytic solutions to the OMT problem are available. To further quantify the quality of our results, we compare them to results obtained with two state-of-the-art deep learning-based OMT solvers: ~\cite{OTICNN} and \cite{korotin2019wasserstein}. We choose these two specific algorithms among the available plethora of available OMT solvers based on the recent benchmark paper by~\cite{korotin2021neural}.
Since both of these algorithms are also based on an ICNN structure, we were able to choose the same architecture with same hyperparameters for all three algorithms, thereby ensuring a fair comparison. We want to emphasize that these parameters could be further fine tuned for all the algorithms and specific experiments to improve the results. To demonstrate the scalability of our algorithm by performing the same experiment in  dimensions 2, 3, 5, 8, 15 and 30.

We do not present comparisons of our approach to more traditional OMT algorithms such as the Sinkhorn algorithm~\citep{cuturi2013sinkhorn} or the linear programming approaches~\citep{peyre2019computational}, as these frameworks, although they approximate the OMT distances, do not compute the continuous optimal transport map, which is essential for the proposed density estimation. While finite element or finite difference based Monge-Ampere solvers, see e.g.~\citep{benamou2019minimal,jacobs2020fast,benamou2000computational}, calculate the continuous OMT map, they are not suitable in dimensions greater than two or three.

To quantify the quality of an estimated transport plan $T$, we present two quantities: the percentage error between the analytic Wasserstein distance and the approximated distance and the $\mathcal{L}^2$-UVP unexplained variance percentage (UVP), which is given by 
\begin{equation}\label{eq:uvp}
    \mathcal{L}^2\mbox{-}\operatorname{UVP}(T) = 100 \cdot \|T - T^*\|^2_{\mathcal{L}^2(\mu)}/\operatorname{Var}(\nu), 
\end{equation}
where $T^*$ denotes the (analytic) optimal  transport plan.

Our first series of experiments is the same as in the benchmark paper~\citep{korotin2021neural}:
we use the gradient of a random convex function to transport the unit Gaussian to a random density. By Brenier's Theorem, as the optimal transport map is the gradient of a convex function, this map is the optimal transport map. In each dimension, we repeated this experiment 20 times to compute the error statistics, which are presented in Table~\ref{tbl:L2UVP}. Wheras all three algorithms seem to work well for this experiment, PICANNs consistently outperform the other algorithms.  As already observed in \cite{korotin2021neural}, this experiment is favoring the ICNN architecture, as the true solution was chosen to be of the same nature, which explains the nearly perfect performance of all three algorithms. Next, we turn to cases where the analytical solution is known in closed form. The first is the special case where both densities are from a family of Gaussians distributions. In that case, the OMT map is simply given by an affine transform and the OMT distance is again given in closed form. We again repeat this experiment 20 times for each dimension,
where we generate Gaussian distributions with a random mean and covariances. Here the means are sampled from a uniform distribution on $[-1,1]$. To construct the random covariance matrices, we recall that we need to enforce the matrix to be positive definite and symmetric. Therefore, we generate a random matrix $A$ of dimension $d\times 3d$, where $d$ is the dimension of the space and where the entries are i.i.d. chosen from a uniform distribution on $[0,0.75]$. Then, a random covariance matrix can be constructed by letting $\Sigma=AA^T$ (the particular form of $\Sigma$ almost surely guarantees positive definiteness). 
The results and comparisons with the other two methods are again presented in Table~\ref{tbl:L2UVP}.  In general, all three algorithms still lead to a good approximation, where one can see that W2Gen and PICANNs are performing significantly better than OT-ICNN. 

The experiments so far do not utilize any complex densities as target distributions. To further validate our algorithm, we choose a more challenging problem: an annulus density for which we know the transport map in closed form. The annulus distribution is given by a push forward of the Gaussian distribution by a gradient of a radially symmetric convex function. This distribution is  given by
$f = g((X^TX)X)\cdot 3(X^TX)^d$ where $g$ is the unit Gaussian and $d$ is the number of dimensions. One can easily check  that the transport map $X \mapsto (X^T X)X$ is the gradient of the convex function $\frac{1}{4} (X^T X)^2$. Thus, we again have access to the optimal transport map; see Figure \ref{fig:Annulus_Results} for a visualization in dimension two. In this figure and in Table \ref{tbl:L2UVP} one can see that PICANNs outperforms both other algorithms by orders of magnitudes.


\begin{table}[htbp]
    \centering
    \begin{tabular}{|c|c|c|c|c|c|c|c|}
    \hline
        \multicolumn{8}{|c|}{\textbf{$\mathcal{L}^2$- Unexplained  Variance  Percentage ($\mathcal{L}^2$-UVP)}} \Tstrut\Bstrut \\
        \hline
        \multirow{3}{*}{\textbf{Experiment}} & \multirow{3}{*}{\textbf{Method}} & \multicolumn{6}{c|}{\textbf{Dimensions}} \TstrutS\\
        \cline{3-8} 
        & & 2d & 3d & 5d & 8d & 15d & 30d \TstrutS\\
        \hline
       \multirow{3}{*}{
       \begin{minipage}[t]{0.23\columnwidth}
       \centering
       Random Cvx Function
       \end{minipage}
       } & PICANNs & \textbf{0.004} & \textbf{0.007} &\textbf{ 0.021} & \textbf{0.034} & \textbf{0.144} & \textbf{0.38 } \TstrutS \\
        & OT-ICNN & 0.043 & 0.052 & 0.145 & 0.276 & 0.746 & 3.98\\
        & W2GEN & 0.040 & 0.043 & 0.046 & 0.052 & 0.150  & 0.60\\
        \hhline{|=|=|=|=|=|=|=|=|}
         \multirow{3}{*}{
           \begin{minipage}[t]{0.23\columnwidth}
            \centering
              Random Gaussian
           \end{minipage}
        } & PICANNs & 0.33  & 0.15 & \textbf{0.15}  & 0.28 & \textbf{0.30}  & 1.13 \TstrutS \\
        & OT-ICNN  & 0.28  & 0.71 & 0.86  & 2.38 & 2.84 & 2.24 \\
        & W2GEN & \textbf{0.17} & \textbf{0.14} & 0.16 & \textbf{0.23} & 0.37 & \textbf{0.67} \\
        \hhline{|=|=|=|=|=|=|=|=|}
       \multirow{3}{*}{
        \begin{minipage}[t]{0.23\columnwidth}
        \centering
        Annulus
        \end{minipage}
       } & PICANNs & \textbf{0.29}  & \textbf{0.43}  & \textbf{0.63 } & \textbf{1.61} & \textbf{7.53} & \textbf{21.71} \TstrutS \\
        & OT-ICNN  & 23.84 & 9.98 & 28.21 & 43.87 & 44.52 & 2725.15 \\
        & W2GEN  & 1.33 & 6.86 & 18.31 & 20.50 & 23.28 & 34.19 \\
    \hline
        \multicolumn{8}{|c|}{\textbf{Avg \% error between true and approximated Wasserstein distance}} \Tstrut\Bstrut \\\hline
       \multirow{3}{*}{
       \begin{minipage}[t]{0.23\columnwidth}
       \centering
       Random Cvx Function
       \end{minipage}
       } & PICANNs & 0.12 & \textbf{0.05} & \textbf{0.03} & \textbf{0.03}  &\textbf{ 0.02} & \textbf{0.04}  \TstrutS\\
        & OT-ICNN & 0.10 & 0.10 & 0.08 & 0.07  & 0.10 & 0.09 \\
        & W2GEN &  \textbf{0.09}  & 0.067  & \textbf{0.03} & 0.04 & 0.06 & 0.52 \\
        \hhline{|=|=|=|=|=|=|=|=|}
         \multirow{3}{*}{
           \begin{minipage}[t]{0.23\columnwidth}
           \centering
              Random Gaussian
           \end{minipage}
        } & PICANNs & \textbf{1.56} & 0.88 & \textbf{0.35} & \textbf{0.21} & \textbf{0.19} & \textbf{0.15} \TstrutS \\
        & OT-ICNN & 1.66 & 1.40  & 0.93 & 0.95 & 0.27 & 0.31 \\
        & W2GEN & 1.59 & \textbf{0.75} & 0.41 & 0.25 & 0.35 & 0.19 \\         \hhline{|=|=|=|=|=|=|=|=|}
       \multirow{3}{*}{
        \begin{minipage}[t]{0.23\columnwidth}
        \centering
        Annulus
        \end{minipage}
       } & PICANNs & \textbf{5.37}  & \textbf{1.25}  & \textbf{1.81}  & \textbf{1.56}  & 7.44 & 5.07  \TstrutS \\
        & OT-ICNN & 6.54 & 25.89& 33.50 & 20.88 & 2.03 & 36.13 \\
        & W2GEN & 12.36 & 19.39 & 8.10 & 3.34 & \textbf{0.96} & \textbf{0.66}  \\
        \hline
    \end{tabular}
    \caption{ We present a comparison between our PICANN approach and \cite{OTICNN} and \cite{korotin2019wasserstein}. In this table, we present the L2-UVP and the percentage error between the theoretical W2 metric and the approximated W2 metric for three experiments. In all these experiments, the source density is the unit Gaussian. The target density in the case of the "Random Cvx Function" experiment is the unit Gaussian deformed by the gradient of a random convex function. In the case "Random Gaussian", the target density is another Gaussian with a randomly sampled mean and co-variance matrix. In the third experiment, the target density is the annulus distribution. The results in the first two experiments are averages of 20 realizations.}
    \label{tbl:L2UVP}
\end{table}

\subsection{Density Estimation Examples}

\begin{figure*}[htbp]
    \centering
    \begin{subfigure}[t]{0.32\textwidth}
        \centering
        \includegraphics[width=\textwidth]{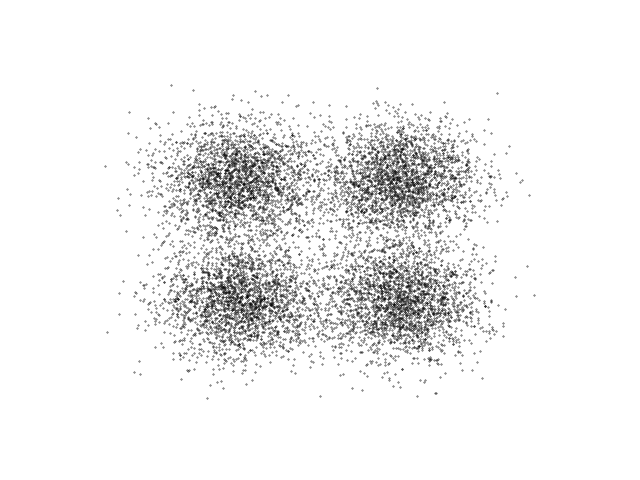}
        \caption{Data}
    \end{subfigure}
    \begin{subfigure}[t]{0.32\textwidth}
        \centering
        \includegraphics[width=\textwidth]{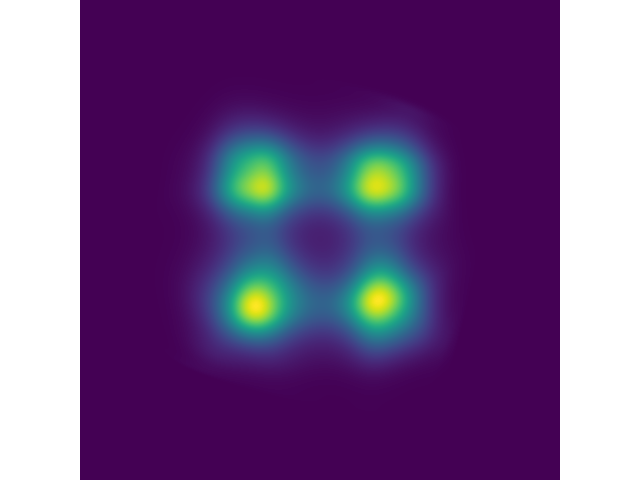}
        \caption{Est. density}
        \label{subfig:GM_approx}
    \end{subfigure}
    \begin{subfigure}[t]{0.32\textwidth}
        \centering
        \includegraphics[width=\textwidth]{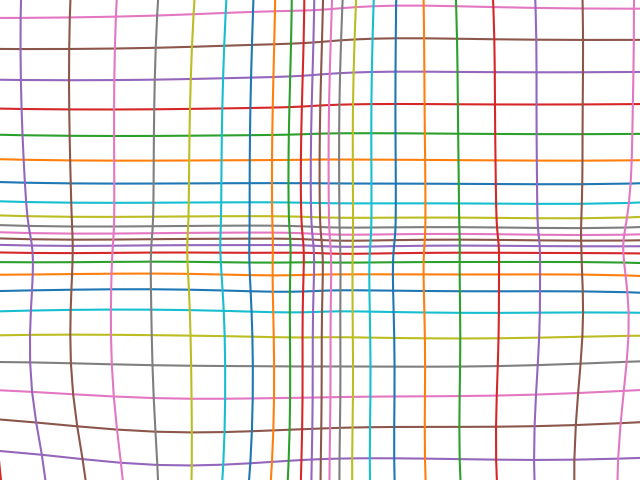}
        \caption{Est. map}
        \label{subfig:GM_Deformation}
    \end{subfigure}
    \caption{Density Estimation 1: In this figure we show an example for density estimation using a simple Gaussian mixture. Panel (a) shows the given data; the approximated density and the inverse map as found using our PICANN approach are shown in Panels (b) and (c).}
    \label{fig:Gaussian_Mixture}
\end{figure*}

\begin{figure*}[htbp]
    \centering
    \begin{subfigure}[t]{0.32\textwidth}
        \centering
        \includegraphics[width=\textwidth]{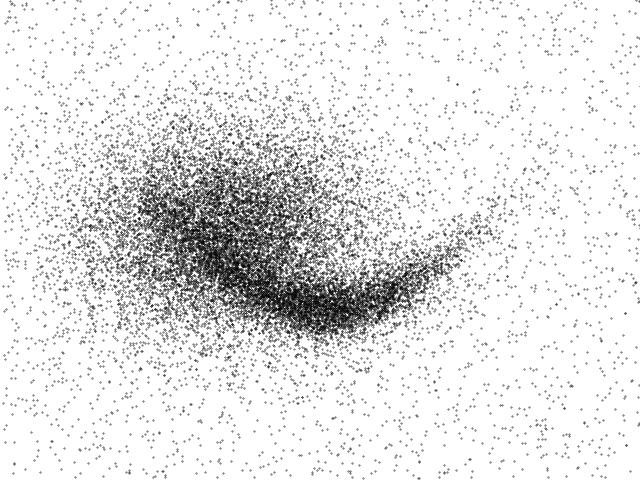}
        \caption{Data}
        \label{subfig:Funny_Samples}
    \end{subfigure}
    \begin{subfigure}[t]{0.32\textwidth}
        \centering
        \includegraphics[width=\textwidth]{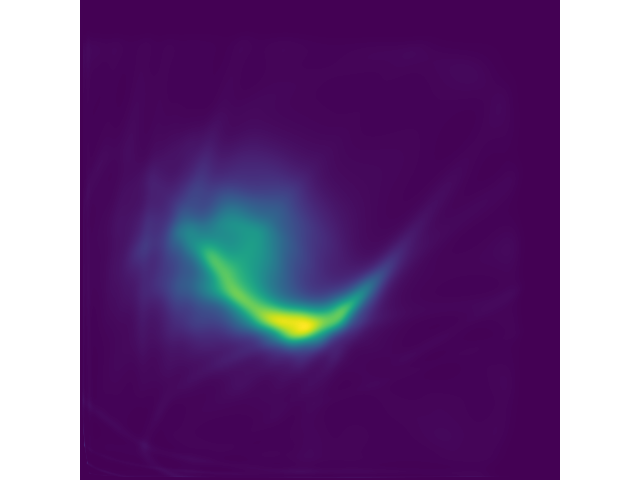}
        \caption{Est. density}
        \label{subfig:Funny_Approx}
    \end{subfigure}
    \begin{subfigure}[t]{0.32\textwidth}
        \centering
        \includegraphics[width=\textwidth]{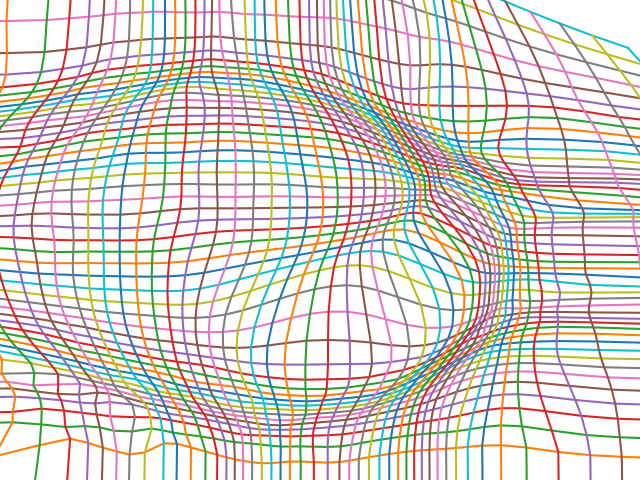}
        \caption{Est. map}
        \label{subfig:Funny_Deformation}
    \end{subfigure}
    \caption{Density Estimation 2: In this figure we show a second example for density estimation. Panel (a) shows the given data; the approximated density and the inverse map as found using our PICANN approach are shown in Panels (b) and (c).}
    \label{fig:FunnyDist}
\end{figure*}

In this section, we present the results for using  the PICANN framework for density estimation. We consider the problem of estimating a continuous density from discrete finite samples. Shown in Figure~\ref{fig:Gaussian_Mixture} are 10k random samples generated from a known Gaussian mixture model of $4$ Gaussians. We use the standard normal distribution as the reference distribution and estimate the optimal transport map between the data and the reference using our PICANN approach. The pushforward of the reference distribution by the estimated transport map and the estimated transport map are both shown in Figure~\ref{fig:Gaussian_Mixture}. We can see that the estimated density matches the original Gaussian mixture. 

Next, we consider a more challenging example: Figure~\ref{fig:FunnyDist}, shows 20k random samples from an nonsymmetric distribution that has been constructed in~\cite{bauer2017diffeomorphic}. We again present the estimated density and transport map. Similar as in the first example, we obtain a good match with the original distribution, where one needs a highly nonlinear transport map.

\section{Conclusion}
In this paper, we use the $L^2$-Wasserstein metric and optimal mass transport (OMT) theory to formulate a density estimation estimation and generative modeling framework. We develop a new deep learning-based solver for the continuous OMT problem, which is rooted in Brenier's celebrated theorem. This theorem allows us to formulate the density estimation problem as a solution to a nonlinear PDE -- a Monge-Ampere equation. Recent developments in deep learning for PDEs, namely PINNS and ICNNs, allow us to develop an efficient solver. We demonstrate the accuracy of our framework by comparing our results to analytic Wasserstein distances. To further quantify the quality of our results we compare them to the results obtained with the two best performing algorithms of the recent benchmark paper for deep learning based OMT solvers~\citep{korotin2021neural}. Our experiments show that our approach significantly outperforms these methods in term of accuracy.
Finally, we present examples of diffeomorphic density estimation within our framework and, in the appendix, we showcase an example of a generative model.

\bibliography{references}
\bibliographystyle{iclr2022_conference}
\appendix
\section{Generative Modeling  with PICANNs} 
\begin{figure*}[htbp]
    \centering
    \includegraphics[width=\textwidth]{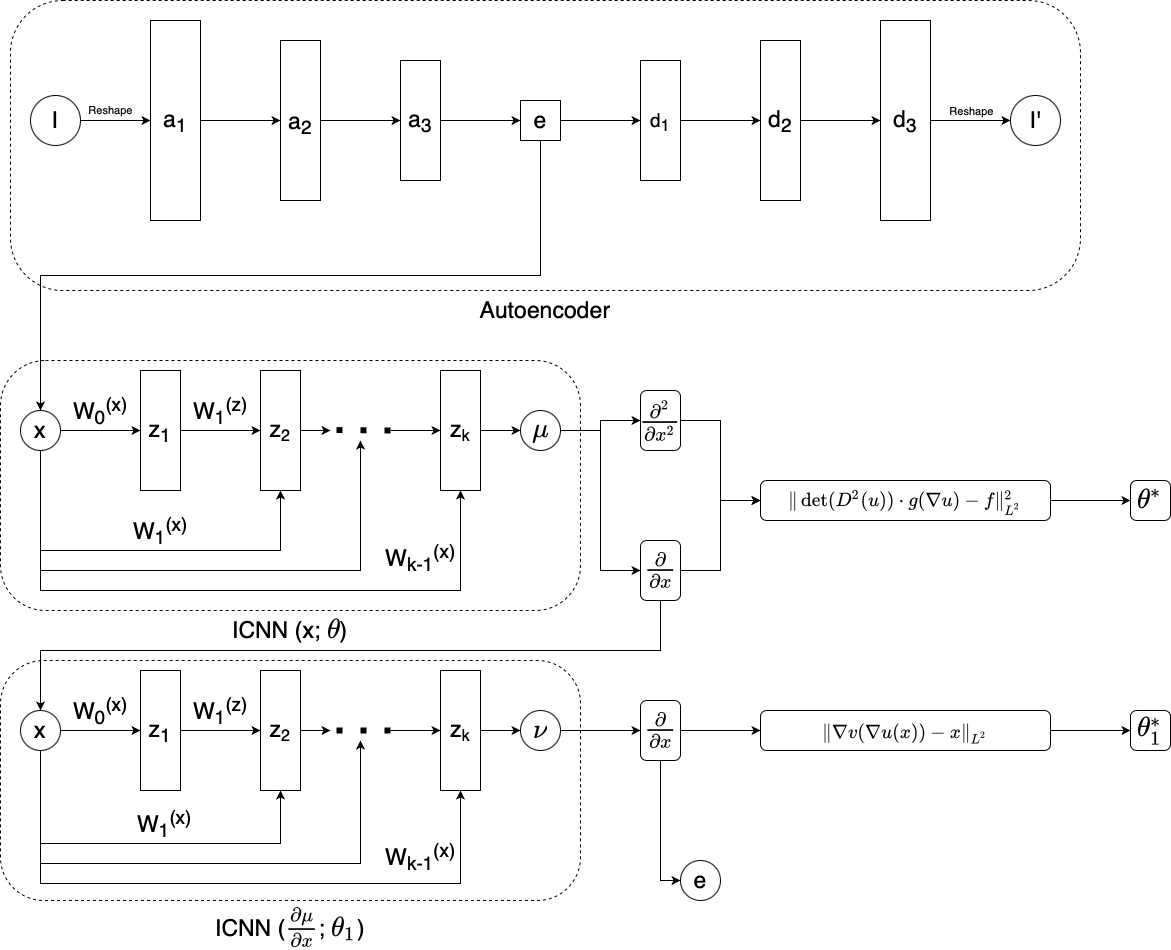}
    \caption{How a combination of an autoencoder with our PICANN approach can be used to develop a generative model. Note how the latent space of the autoencoder becomes the input to the PICANN network. In such a setting, the PICANN estimates the latent space density and samples from the estimated distribution. Using random samples from this distribution, one can pass them through the decoder to generate new samples.}
    \label{fig:generative_autoen}
\end{figure*}

In this appendix, we present an application of the PICANN approach to develop a generative model.  For this task, we operate within the  autoencoder setting, which is essentially a dimensionality reduction technique~\citep{kramer1991nonlinear}. The autoencoder maps high-dimensional data to a latent space of lower dimensions, which then can be transformed back to the original space using the "decoder" part of the network. Fig~\ref{fig:generative_autoen}, summarizes how the autoencoder can be naturally included in our PICANN approach to obtain an efficient generative model for high-dimensional data.

\begin{figure*}[htbp]
    \centering
    \begin{subfigure}[!ht]{0.45\textwidth}
        \centering
        \includegraphics[height=1.6in]{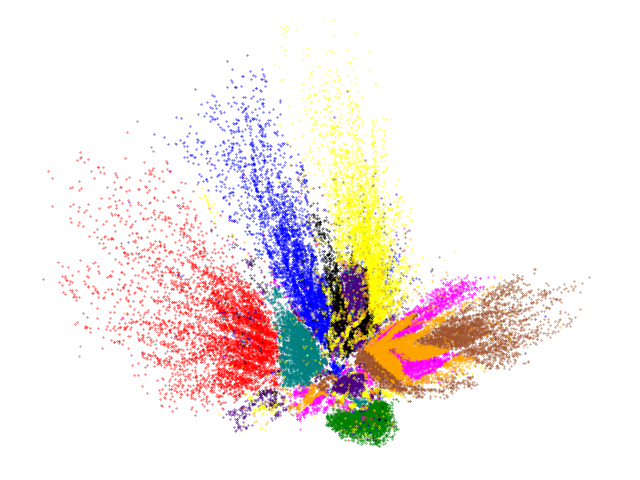}
        \caption{Encoded MNIST}
    \end{subfigure}
     \begin{subfigure}[!ht]{0.45\textwidth}
        \centering
        \includegraphics[height=1.6in]{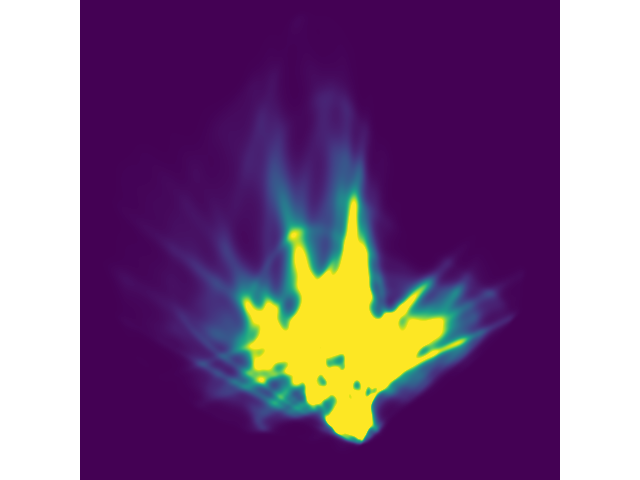}
        \caption{Est. density}
        \label{subfig:estimated_density}
    \end{subfigure}
    \vspace{.1cm}
    
    \begin{subfigure}[!ht]{0.45\textwidth}
        \centering
       \includegraphics[height=1.6in]{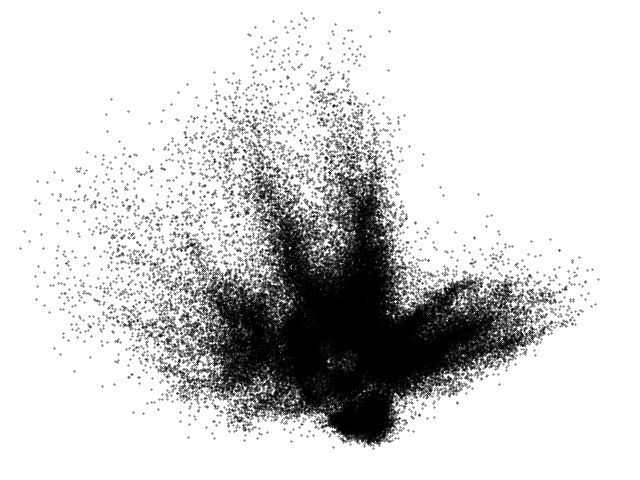}
        \caption{Samples}
        \label{subfig:MNIST_LatentSamples}
    \end{subfigure} 
    \begin{subfigure}[!ht]{0.45\textwidth}
        \centering
        \includegraphics[height=1.6in]{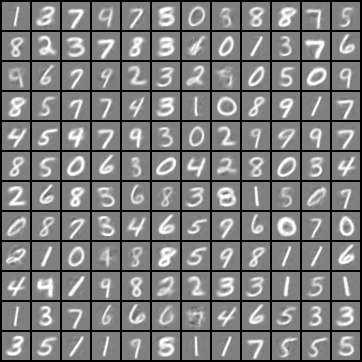}
        \caption{Output}
        \label{subfig:Decoded_Random_Samples}
    \end{subfigure}
    \caption{Generative model: this figure details the application of our framework to the  MNIST dataset.  
    Panel (a) shows the encoded samples from the MNIST database.
    A PICANN was trained to estimate this distribution and learn the forward and inverse transport map between the encoded 'MNIST Distribution' and a  Gaussian. The estimated
    density can be seen in Panel~(b), and Panel~(c) shows 100k samples  generated from this density using our obtained optimal transport map. The first 144 samples passed through the decoder part of the autoencoder are shown in Panel (d).}
    \label{fig:MNIST}
\end{figure*}

To demonstrate the effectiveness of the generative algorithm, we consider the MNIST dataset encoded using a simple fully connected autoencoder to a latent space of dimension 2. Figure~\ref{fig:MNIST} shows the encoded data points with each class assigned a unique color. We train a PICANN network of 5 hidden layers with 128 neurons in each layer, to learn the forward and the inverse mapping from a Gaussian with the mean and covariance of the encoded data to the unknown "latent MNIST distribution". Figure~\ref{fig:MNIST} panel(c)  presents the samples as generated from the Gaussian and then transported points using the approximated inverse map. Panel (d) displays the result of passing a random subset of these generated samples through the decoder.

\end{document}

%% file: math_commands.tex
\usepackage{amsmath,amsfonts,bm}

\def\eqref#1{equation~\ref{#1}}

\def\1{\bm{1}}

\DeclareMathAlphabet{\mathsfit}{\encodingdefault}{\sfdefault}{m}{sl}
\SetMathAlphabet{\mathsfit}{bold}{\encodingdefault}{\sfdefault}{bx}{n}

